\pdfoutput=1

\documentclass[11pt]{article}
\PassOptionsToPackage{table}{xcolor}
\usepackage{acl}
\usepackage[normalem]{ulem}
 \usepackage{amsmath} 
\usepackage{times}
\usepackage{latexsym}
\usepackage{amsfonts}
\usepackage{booktabs} % 提供更美观的表格线条
\usepackage{multirow} % 支持合并行
\usepackage{graphicx} % 支持缩放表格
\usepackage{float} % Allows precise float placement with H
\usepackage[T1]{fontenc}
\usepackage[utf8]{inputenc}

\usepackage{microtype}

\usepackage{inconsolata}

\usepackage{graphicx}
\usepackage{pifont}

\usepackage{soul}

\definecolor{DarkGreen}{RGB}{9, 136, 66} %
\definecolor{Maroon}{RGB}{238, 49, 49}    %
\newcommand{\greenCheckmarkBold}{\textcolor{DarkGreen}{\ding{51}}} %
\newcommand{\redXSolidBrush}{\textcolor{Maroon}{\ding{55}}}    %

\title{DocFusion: A Unified Framework for Document Parsing Tasks}

\author{
\textbf{Mingxu Chai}\textsuperscript{1,2}\thanks{Equal contribution}, 
\textbf{Ziyu Shen}\textsuperscript{1}\footnotemark[\value{footnote}], 
\textbf{Chong Zhang}\textsuperscript{1}, 
\textbf{Yue Zhang}\textsuperscript{1}, \\
\textbf{Xiao Wang}\textsuperscript{1}, 
\textbf{Shihan Dou}\textsuperscript{1}, 
\textbf{Jihua Kang}\textsuperscript{1}, 
\textbf{Jiazheng Zhang}\textsuperscript{1}, 
\textbf{Qi Zhang}\textsuperscript{1}\thanks{Corresponding author} \\
\textsuperscript{1}Computation and Artificial Intelligence Innovative College, Fudan University, Shanghai, China \\
\textsuperscript{2}Shanghai Innovation Institute, Shanghai, China \\
\texttt{\{qz\}@fudan.edu.cn}
}

\begin{document}

\maketitle

\begin{abstract}

Document parsing involves layout element detection and recognition, essential for extracting information. However, existing methods often employ multiple models for these tasks, leading to increased system complexity and maintenance overhead. While some models attempt to unify detection and recognition, they often fail to address the intrinsic differences in data representations, thereby limiting performance in document processing. Our research reveals that recognition relies on discrete tokens, whereas detection relies on continuous coordinates, leading to challenges in gradient updates and optimization. To bridge this gap, we propose the Gaussian-Kernel Cross-Entropy Loss (GK-CEL), enabling generative frameworks to handle both tasks simultaneously. Building upon GK-CEL, we propose DocFusion, a unified document parsing model with only 0.28B parameters. Additionally, we construct the DocLatex-1.6M dataset to provide high-quality training support. Experimental results show that DocFusion, equipped with GK-CEL, performs competitively across four core document parsing tasks, validating the effectiveness of our unified approach. The model and datasets are publicly available at: \url{https://github.com/sc22mc/DocFusion}
\end{abstract}

\section{Introduction}

Document parsing plays a significant role in extracting structured data from documents, making it foundational for various downstream applications. For example, in Retrieval-Augmented Generation (RAG) workflows \cite{ren2023rocketqav2jointtrainingmethod, zhang2022adversarialretrieverrankerdensetext}, extracting well-organized and contextually rich information from documents can improve the performance of large language models (LLMs) \cite{ zhao2024surveylargelanguagemodels, gao2024retrievalaugmentedgenerationlargelanguage}. However, real-world documents often embed information in complex structures, such as hierarchical layouts, mathematical expressions, and tables, which pose considerable challenges for automated parsing.

\begin{table}[t]
  \centering
  \resizebox{\columnwidth}{!}{
  \begin{tabular}{lcccccc}
    \toprule
    \multicolumn{1}{c}{\textbf{Tool Type}} & \textbf{Size} & \textbf{DLA} & \textbf{MER} & \textbf{TR} & \textbf{OCR} \\
    \bottomrule \addlinespace
    \textbf{System} & & & & &\\
    \hline
    open-parse \citeyearpar{open} &- &\greenCheckmarkBold &\redXSolidBrush &\greenCheckmarkBold &\greenCheckmarkBold\\
    LlamaParse \citeyearpar{llama} &- &\greenCheckmarkBold &\greenCheckmarkBold &\greenCheckmarkBold &\greenCheckmarkBold\\
    DeepDoc \citeyearpar{deepdoc} &- &\greenCheckmarkBold &\redXSolidBrush &\greenCheckmarkBold &\greenCheckmarkBold\\
    MinerU \citeyearpar{wei2024generalocrtheoryocr20} &- &\greenCheckmarkBold &\greenCheckmarkBold &\greenCheckmarkBold &\greenCheckmarkBold\\
    \bottomrule \addlinespace
    \textbf{Model} & & & & &\\
    \hline
    DocYOLO\citeyearpar{zhao2024doclayoutyoloenhancingdocumentlayout} &20M &\greenCheckmarkBold &\redXSolidBrush &\redXSolidBrush &\redXSolidBrush       \\

    ViTLP \citeyearpar{mao2024visually} &253M &\greenCheckmarkBold &\redXSolidBrush &\greenCheckmarkBold &\greenCheckmarkBold\\
    
    UniMER \citeyearpar{wang2024unimernetuniversalnetworkrealworld} &325M &\redXSolidBrush &\greenCheckmarkBold &\redXSolidBrush &\redXSolidBrush       \\
    
    Nougat \citeyearpar{blecher2023nougatneuralopticalunderstanding} &350M &\redXSolidBrush &\greenCheckmarkBold &\greenCheckmarkBold &\greenCheckmarkBold\\
    GOT \citeyearpar{wei2024generalocrtheoryocr20} &580M  &\redXSolidBrush &\greenCheckmarkBold &\greenCheckmarkBold &\greenCheckmarkBold\\
    StructTable \citeyearpar{xia2024docgenome} &938M &\redXSolidBrush &\redXSolidBrush &\greenCheckmarkBold &\redXSolidBrush \\
    \bottomrule \addlinespace
    \rowcolor{gray!30} 
    DocFusion(Ours) &\cellcolor{yellow!50}289M &\greenCheckmarkBold &\greenCheckmarkBold &\greenCheckmarkBold &\greenCheckmarkBold\\
    \bottomrule
  \end{tabular}
  }
  \caption{\label{function}
    Capabilities of document parsing tools. \textbf{Model} refers to a single  model, while \textbf{System} integrates multiple models. \textbf{DLA}: Document Layout Analysis. \textbf{MER}: Math Expression Recognition. \textbf{TR}: Table Recognition. \textbf{OCR}: Optical Character Recognition. Compare with multi-model systems, DocFusion achieves all four tasks within a single model, requiring only 289M parameters.
}
\end{table}

Existing methods can be categorized into two main approaches: multi-module pipeline systems and end-to-end page-level OCR models. Multi-module pipeline systems decompose document parsing tasks into independent modules, allowing each module to adopt the best model. For example, DocLayout-YOLO \cite{zhao2024doclayoutyoloenhancingdocumentlayout} has demonstrated excellent performance in Layout analysis, while UniMERNet \cite{wang2024unimernetuniversalnetworkrealworld} achieves SOTA results in Math Expression Recognition. Although this approach improves performance for specific tasks, integrating multiple models into a single system increases overall complexity. Moreover, these systems fail to fully exploit task-level collaboration, leading to inefficiencies in parameter usage. In contrast, end-to-end page-level OCR models, such as Nougat \cite{blecher2023nougatneuralopticalunderstanding} and GOT \cite{wei2024generalocrtheoryocr20}, can seamlessly integrate multiple recognition tasks. While the outputs of these models demonstrate a well-organized logical structure, the models lack the ability to generate bounding boxes for layout elements. As a result, they fail to preserve the spatial relationships between documents and their layouts, which is crucial for interpretability in RAG workflows. Additionally, while these models perform well on page-level images, it struggles with specific layout elements,  limiting their flexibility in application. To address these issues, this research focused on four key tasks: document layout analysis (\textbf{DLA}), mathematical expression recognition \textbf{(MER)}, table recognition \textbf{(TR)}, and optical character recognition \textbf{(OCR)}.

Several studies have attempted to apply generative frameworks to integrate object detection and content recognition, achieving promising results on natural images \cite{xiao2311florence}. However, extending such frameworks to document images presents significant challenges due to the inherent structural and representational differences between these domains. Through experiments, we identify the primary issue as the fundamental conflict between the continuous nature of coordinate data and the discrete nature of token generation, which disrupts gradient updates during multi-task training (discussed in Section \ref{3.2}). In natural scene images, small deviations in coordinates and text are generally tolerable. However, in document parsing, even minor errors in LaTeX code can critically impact compilation success rates. This imposes stricter accuracy requirements when applying such frameworks to document understanding tasks. To address these challenges, we propose Gaussian-Kernel Cross-Entropy Loss (GK-CEL), an improved objective function designed to mitigate the inconsistencies between discrete and continuous data representations, enhancing the performance of generative frameworks in document parsing.

MER and TR are essential for LaTeX-based document processing, but existing datasets suffer from inconsistent formatting and redundant characters, where different writing styles generate identical compiled outputs, introducing noise that hinders model training (details are provided in Appendix \ref{appendixMERandTR}). To address this, we propose DocLatex-1.6M, a large-scale, high-quality dataset that enhances annotation consistency and improves model training efficiency. Experiments demonstrate that DocFusion trained on this dataset outperforms task-specific models with fewer parameters.

Our contributions are summarized as follows:

\begin{itemize}
    \item We propose DocFusion, a unified generative multi-task model that standardizes task formulations and achieves SOTA performance across four key document parsing tasks.
    
    \item We propose GK-CEL to resolve the conflict between continuous coordinate and discrete token in the generative framework, enhancing document parsing and offering a reference for similar frameworks in other domains.
    
    \item Experimental results demonstrate that incorporating multi-task data significantly outperforms single-task setups, providing insights into the benefits of multi-task learning in document parsing.
    
    \item We constructed DocLatex-1.6M, a large-scale, high-quality dataset with 1.5M LaTeX-annotated math expressions and 100K tables, offering a valuable resource for advancing document parsing research.
\end{itemize}

% 1.
\begin{figure*}[t]
  \includegraphics[width=0.96\linewidth]{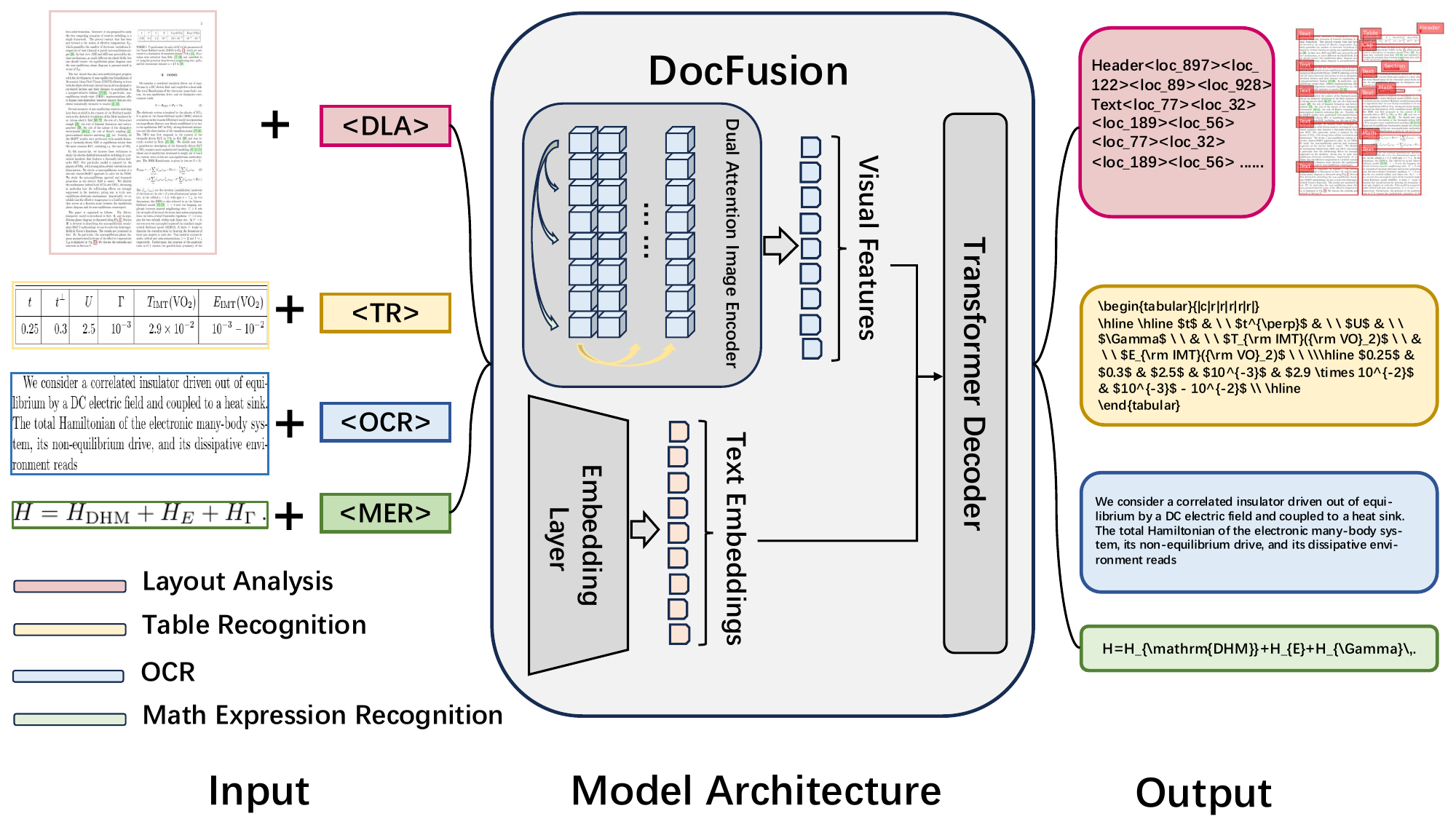}
  \caption {The model comprises three key components: a visual encoder, a text embedding layer and a Transformer decoder. The image features extracted by the visual encoder and the instruction embeddings are combined and then passed to the Transformer decoder, which produces the final output sequence.}
    \label{2}
\end{figure*}

\section{Related Work}

\textbf{Document Parsing Models.} Document parsing models have seen  remarkable progress across various tasks. DLA has evolved from vision-based methods \cite{Wick_Puppe_2018,Bao_Dong_Wei_2021} to multimodal approaches integrating textual features \cite{xu2022layoutlmv2multimodalpretrainingvisuallyrich,huang2204layoutlmv3}. OCR has transitioned from template matching \cite{4376991} to deep learning-based solutions \cite{8237504,Chen_Li_Xue_2021,10476585}. MER progressed from symbol segmentation \cite{miller1998ambiguity} to CNN-RNN hybrids \cite{le2019pattern} and Transformer-based models \cite{wang2024unimernetuniversalnetworkrealworld}. Similarly, TR now employs methods like grid segmentation and image-to-sequence techniques to reconstruct structured data \cite{qasim2019rethinking,huang2023improving,xia2024docgenome}. Page-level end-to-end OCR models like Nougat \cite{blecher2023nougatneuralopticalunderstanding} and GOT \cite{wei2024generalocrtheoryocr20} simplify workflows by integrating multi recognition tasks.\\

\textbf{Modular Pipeline Systems.} The advancements in task-specific models have driven the development of modular pipeline systems, which process complex document structures through specialized modules. For instance, Open-Parse\cite{open} performs well in incrementally parsing complex layouts but lacks support for MER. Other systems, such as DeepDoc\cite{deepdoc} and Llama-Parse\cite{llama}, extend the scope of modular pipelines to handle more diverse tasks. In particular, MinerU\cite{MinerU} stands out by supporting advanced features such as complex layout parsing and Markdown conversion.

\section{Method}
% We propose DocFusion, a unified framework for document parsing. 
We introduce the model architecture (\ref{3.1}) and explain how detection tasks are represented into the generative framework. Then, we discuss the challenges (\ref{3.2}) of detection tasks within this framework. Next, we explain the Gaussian-Kernel Cross-Entropy Loss(\ref{3.3})
% , which facilitates efficient joint training.

\subsection{Architecture}
\label{3.1}
As shown in Figure \ref{2}, the architecture of DocFusion consists of three main components: a vision encoder, a text embedding layer, and a Transformer decoder. Since the task instructions are limited and predefined, no Transformer encoder is included, task-specific prompts are directly embedded, simplifying the architecture. 

To unify the representation of object detection and text recognition tasks, we adopt a coordinate quantization representation \cite{xiao2311florence}. Specifically, images are quantized into a fixed resolution (e.g., 1000×1000), and coordinates are represented as discrete tokens (e.g., <loc\_1>, <loc\_2>, ..., <loc\_1000>). This approach enables the use of a unified generative framework for detection tasks. To address the challenges posed by densely structured content, the vision encoder incorporates a Dual Attention mechanism \cite{Ding_2_Codella_1_Wang_Yuan_Kong_Cloud}, which captures interactions across channel and spatial dimensions, enhancing feature extraction for intricate document layouts. Additionally, the traditional feed-forward network (FFN) is removed, reducing both parameter count and computational cost, further improving model efficiency.

The vision encoder processes input images $\mathbf{I} \in \mathbb{R}^{H \times W \times 3}$ into visual features, flattened as token embeddings $\mathbf{V} \in \mathbb{R}^{N_v \times D_v}$. These embeddings are projected to $D_t$, resulting in $\mathbf{V'} \in \mathbb{R}^{N_v \times D_t}$, to match the task-specific prompt embeddings $\mathbf{T}{\text{prompt}} \in \mathbb{R}^{N_t \times D_t}$. The combined input $\mathbf{X} = [\mathbf{V'}; \mathbf{T}{\text{prompt}}]$ is then passed to the Transformer decoder to generate predictions. 

\begin{figure}[tbp]
  \includegraphics[width=\columnwidth, trim=260 110 260 102, clip]{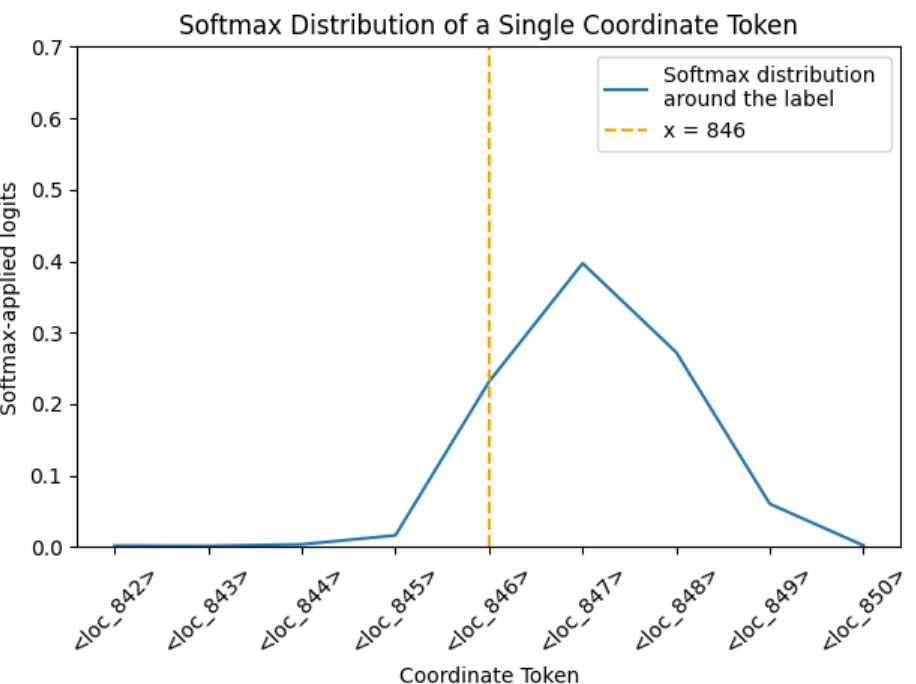}
  \caption{The distribution of logits for a target token after the loss has stabilized when using the Common CE Loss.}
  \label{softmax_dis}
\end{figure} 

\subsection{Challenges and Motivations}
\label{3.2}
While representing object detection as text generation enables joint training of layout analysis and page element recognition under a unified cross-entropy-based framework, it inherently forces continuous coordinates into discrete token spaces. This mismatch creates several challenges, especially in fine-tuning small coordinate adjustments, where the model struggles to produce accurate gradients, reducing training stability. As shown in Figure \ref{softmax_dis}, small unavoidable deviations in coordinate labels smooth out the softmax distribution, preventing the target token’s probability from forming a sharp peak. This makes it harder for the model to escape local optima and limits its learning capacity. Additionally, traditional cross-entropy loss, which is designed for discrete classification tasks, does not handle continuous changes well, further increasing inaccuracies during training.

In multi-task settings, these issues become even more challenging. The conflict between discrete loss functions and continuous coordinate optimization can skew gradients, causing one task to dominate at the cost of others. This imbalance reduces performance in other tasks and harms the model’s ability to predict coordinates accurately, limiting its overall effectiveness in complex document parsing tasks. Solving these problems is critical to improving both localization accuracy and training stability across tasks.

\subsection{Gaussian-Kernel Design}
\label{3.3}
\begin{figure}[tbp]
  \includegraphics[width=\columnwidth]{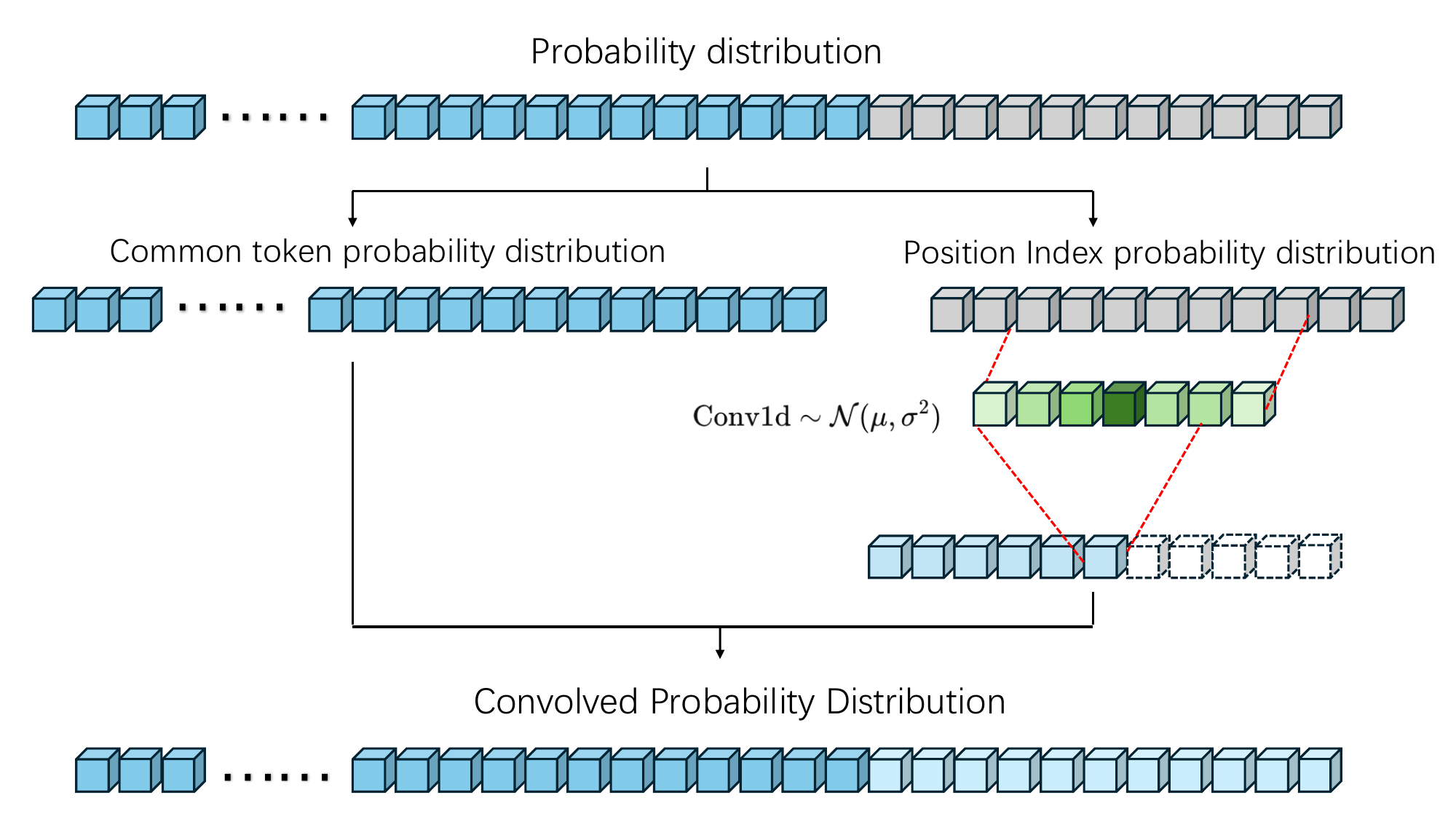}
  \caption{Illustration of Gaussian-Kernel Cross-Entropy Loss.}
  \label{GKCEL_fig}
\end{figure}

To address these challenges, we propose the Gaussian-Kernel Cross-Entropy Loss (GK-CEL). As shown in Figure \ref{GKCEL_fig}, it applies a one-dimensional convolution with Gaussian-distributed weights over the probability distribution, fine-tuning the model’s sensitivity to small coordinate changes while preserving the discrete treatment of cross-entropy. This approach alleviates the mismatch between discrete tokens and continuous coordinates, improves gradient quality, and prevents the coordinate prediction task from dominating the optimization process. As a result, it enhances localization accuracy and supports stable multi-task training.

Let the model's output logits be denoted as $\mathbf{Z} \in \mathbb{R}^{B \times L \times V}$, where $B$ is the batch size, $L$ is the sequence length, and $V$ is the vocabulary size. The target labels are denoted as $\mathbf{T} \in \mathbb{N}^{B \times L}$. The range of indices corresponding to coordinate tokens is defined as $[s, e]$, representing their positions in the vocabulary.

The standard softmax probability distribution is first computed as:
\begin{align}
    \mathbf{P} = \text{softmax}(\mathbf{Z})
\end{align}

A mask is then applied to zero out probabilities outside the range $[s, e]$, creating a modified probability tensor $\mathbf{P}'$:

\begin{align}
\mathbf{P}'_{ijk} = 
\begin{cases}
\mathbf{P}_{ijk}, & \text{if } k \in [s, e] \\
0, & \text{otherwise}
\end{cases}
\end{align}

where $i$ represents the batch index, $j$ represents the sequence position, and $k$ represents the vocabulary index.

Next, a one-dimensional convolution kernel $\mathbf{K} \in \mathbb{R}^{1 \times 1 \times n}$ is constructed based on a Gaussian distribution, where $n$ is the kernel size (an odd integer greater than 1), $\sigma$ is the standard deviation and $p$ represents the position of each element in the convolution kernel, measured as the offset from the center, where the center is located at $\frac{n+1}{2}$. The range of $p \in [1,n]$. The kernel weights of each index are computed as: 
\begin{align} 
    \mathbf{K}_{p} = \exp\left(-\frac{(p - \frac{n+1}{2})^2}{2\sigma^2}\right) 
\end{align}

The kernel is then applied to $\mathbf{P}'$ via one-dimensional convolution:
\begin{align}
    \mathbf{C} = \text{conv1d}(\mathbf{P}', \mathbf{K})
\end{align}

The convolution result $\mathbf{C}$ is integrated back into the original probability distribution $\mathbf{P}$ within the index range $[s, e]$, while retaining the original probabilities outside this range:

\begin{align}
\mathbf{P}''_{ijk} = 
\begin{cases}
\mathbf{C}_{ijk}, & \text{if } k \in [s, e] \\
\mathbf{P}_{ijk}, & \text{otherwise}
\end{cases}
\end{align}

The final objective function is computed as:
\begin{align}
\mathcal{L} = -\frac{1}{N} \sum_{i=1}^{B} \sum_{j=1}^{L} \mathbf{M}_{ij} \log \mathbf{P}''_{ij\mathbf{T}_{ij}}
\end{align}
where $\mathbf{M}_{ij}$ is a mask matrix that indicates whether the target label at position $(i, j)$ should contribute to the loss calculation.The normalization factor $\mathbf{N}$ is defined as the total number of valid targets.

\section{Experiments}
\subsection{Training Datasets}
\label{training datasets}
In the training phase, the DLA task uses the DocLayNet \cite{doclaynet2022} dataset, which contains 80,863 pages from 7 document types and is manually annotated with 11 categories. The images are split into 69,103/6,481/4,999 for training/validation/testing, respectively. The OCR dataset is also sourced from DocLayNet, which offers comprehensive annotations for layout elements and their corresponding text, and is widely regarded as a reliable resource in the academic community. For the TR and MER tasks, we used the DocLatex-1.6M dataset, which was constructed in this work. Additionally, although this work primarily focuses on document images, we introduced the HME100K\cite{yuan2022syntax}, a handwritten math expression dataset to enhance the generalization ability of the MER task.

\begin{table*}[ht]
  \vspace{-0.5cm}
  \centering
  \setlength{\tabcolsep}{5.7pt}
  % \small
  \begin{tabular}{lcccccccc}
  \toprule
    \multirow[c]{2}{*}{\textbf{Model}} & \multirow[c]{2}{*}{\textbf{size}}&\multicolumn{2}{c}{\textbf{OCR}} & \multicolumn{3}{c}{\textbf{MER}}&\multicolumn{2}{c}{\textbf{TR}}  \\
    \cmidrule(lr){3-4} \cmidrule(lr){5-7} \cmidrule(lr){8-9}
& &BLEU$\uparrow$&EditDis$\downarrow$&CDM$\uparrow$&ExpRate$\uparrow$ &CSR$\uparrow$ &F1$\uparrow$ &CSR$\uparrow$\\
    \hline
    % Qwen-VL-PLUS \citeyearpar{bai2023qwen} &- &87.2 &7.8&-&-&-&-&-\\
    UReader \citeyearpar{ye2023ureaderuniversalocrfreevisuallysituated} &7B &38.6 &47.3&-&-&-&-&-\\
    LLaVA-NeXT \citeyearpar{li2024llavanextinterleavetacklingmultiimagevideo} &34B &69.1 &27.2&-&-&-&-&-\\
    Nougat \citeyearpar{blecher2023nougatneuralopticalunderstanding} &250M &71.6 &21.4&-&-&-&-&-\\
    TextMonkey \citeyearpar{liu2024textmonkeyocrfreelargemultimodal} &7B &73.3 &21.9&-&-&-&-&-\\
    Qwen-VL-MAX \citeyearpar{bai2023qwen} &>72B &94.7 &3.9&-&-&-&-&-\\
    Qwen-VL-OCR \citeyearpar{bai2023qwen} &- &95.9 &4.1&-&-&-&-&-\\
    % TextMonkey \citeyearpar{bai2023qwen} &7B &95.4 &0.043&-&-&-&-&-\\
    Pix2tex \citeyearpar{blecher2022pix2tex} &- &-    &-  &76.5  &41.7  &95.9  &-  &-    \\
    Texify \citeyearpar{paruchuri2023texify} &312M &-    &-  &88.6  &71.7  &97.8  &-  &-    \\
    Mathpix &- &-    &-  &88.9  &79.1  &98.3  &-  &-    \\
    UniMERNet \citeyearpar{wang2024unimernetuniversalnetworkrealworld} &325M  &-    &-  &\textbf{99.0}  &89.5  &99.7  &-  &-   \\
    MixTex \citeyearpar{luo2024mixtexunambiguousrecognitionrely} &85M  &-    &-  &-  &-  &-  &46.2  &27.4   \\
    StructEqTable \citeyearpar{xia2024docgenome} &938M  &-    &-  &-  &-  &-  &90.6  &\textbf{93.2}   \\
    GOT \citeyearpar{wei2024generalocrtheoryocr20} &580M  &96.8    &2.2  &87.7  &67.3  &97.8  &86.9   &81.6    \\
    DocFusion(Ours)  &289M   &\textbf{97.4}    &\textbf{1.8}  &98.7  &\textbf{94.2}  &\textbf{99.8}  &\textbf{92.1}  &92.5     \\
    \toprule
  \end{tabular}
  \caption{\label{main}
    Comparison of DocFusion with other models on three recognition tasks in the document scene. Specifically, DocLaynet\cite{doclaynet2022} was used for OCR, DocGenome\cite{xia2024docgenome} for TR, and UniMER-1M\cite{wang2024unimernetuniversalnetworkrealworld} for MER. More details on the TR experiments can be found in Appendix \ref{TR}. \textit{Note:} Nougat is primarily designed for full-page recognition and tends to underperform on isolated tables or mathematical expressions.
  }
\end{table*}

\begin{table*}[ht]
  \centering
  \setlength{\tabcolsep}{4.2pt}
  \resizebox{\textwidth}{!}{
  \begin{tabular}{lcccccccccc}
  \toprule
    \multirow[c]{2}{*}{\textbf{Model}} &\multirow[c]{2}{*}{\textbf{Size}}      & \multicolumn{3}{c}{\textbf{DocLayNet}} & \multicolumn{3}{c}{\textbf{DocLayNet-Scientific}} & \multirow[c]{2}{*}{FPS$\uparrow$} & \multirow[c]{2}{*}{\st{NMS}} & \multirow[c]{2}{*}{\st{Conf}}\\
    \cmidrule(lr){3-5} \cmidrule(lr){6-8}
     & &Precision$\uparrow$&Recall$\uparrow$&F1$\uparrow$&Precision$\uparrow$&Recall$\uparrow$&F1$\uparrow$\\
     
    \hline

    YOLOv10m \citeyearpar{wang2024yolov10realtimeendtoendobject} &16M &90.1  &86.9  &88.4 &94.3  &94.5  &94.4      &93.6   &\greenCheckmarkBold &\redXSolidBrush    \\

    YOLOv11m \citeyearpar{khanam2024yolov11overviewkeyarchitectural} &20M &90.5  &87.4  &88.9 &95.1  &94.9  &95.0      &\textbf{100.8}   &\redXSolidBrush &\redXSolidBrush    \\
    
    YOLO-DocLayout \citeyearpar{zhao2024doclayoutyoloenhancingdocumentlayout} &20M &90.9  &\textbf{88.2}  &\textbf{89.5}  &95.5  &94.4  &95.0      &55.2   &- &\redXSolidBrush    \\

    \hline
    
    DETR  \citeyearpar{carion2020endtoendobjectdetectiontransformers}    &41M    &84.7  &87.1  &85.8  &92.2  &92.0  &92.1      &17.6  &\greenCheckmarkBold  &\redXSolidBrush \\
    
    DETR-Deformable  \citeyearpar{https://doi.org/10.48550/arxiv.2010.04159}    &41M    &\textbf{91.6}  &87.1  &89.3  &96.2  &95.9  &96.0      &18.8  &\greenCheckmarkBold  &\redXSolidBrush \\

    \hline
    
    DocFusion(Ours)   &289M          &88.9  &87.8  &88.4  &\textbf{96.8}  &\textbf{96.2}  &\textbf{96.4}  &11.4 &\greenCheckmarkBold &\greenCheckmarkBold\\
    \toprule
  \end{tabular}
  }
  \caption{\label{end_dla}
The performance of the models on DLA, where DocLayNet-Scientific refers to the scientific document subset of DocLayNet. \textbf{\st{NMS}} indicates that Non-Maximum Suppression is not required, while \textbf{\st{Conf}} means no confidence adjustment is needed. \textit{Note:} The results of DETR and YOLO-series models in this table are computed at multiple confidence levels, with the highest F1 score selected as the final result. \textit{YOLO-DocLayout builds on YOLOv10, which is NMS-free by design. However, due to structural changes, it is unclear whether it can still be fully considered NMS-free.}
}

\end{table*}

\subsection{Evaluation Metrics}

\subsubsection{Evaluation Metrics for Recognition}
We employ BLEU \cite{Papineni_Roukos_Ward_Zhu_2001} and Edit Distance  \cite{Levenshtein} to evaluate sequences. Additionally, CDM \cite{wang2024cdmreliablemetricfair} and CSR were used to better assess the quality of LaTeX-based outputs.\\
\textbf{BLEU} is used for evaluating generated text, measuring n-gram overlap with reference texts.\\
\textbf{Edit distance} measures the minimum number of operations insertions, deletions, or substitutions required to transform one string into another. \\
\textbf{CSR} refers to the percentage of generated LaTeX outputs that can be successfully compiled into PDF.\\
\textbf{ExpRate} \cite{Li_Yuan_Liang_Liu_Ji_Bai_Liu_Bai} measures the proportion of samples where the predicted text matches the reference text without any errors. \\
\textbf{CDM} evaluates MER by comparing image-rendered expression at the character level with spatial localization, ensuring fairness and accuracy over text-based metrics like BLEU.

\subsubsection{Evaluation Metrics for Detection}
Since DocFusion adopts a novel approach in the DLA task without relying on confidence scores, we did not use the widely adopted Average Precision but instead focus on the following metrics:\\
\textbf{Precision} measures the proportion of correctly identified positive instances among all predicted positives.\\
\textbf{Recall} measures the proportion of correctly identified positive instances among all actual positives.\\
\textbf{F1-score} balances precision and recall, serving as their harmonic mean. \\
% This metric is particularly useful for evaluating the trade-off between precision and recall in the DLA task.\\
\textbf{FPS} measures the number of frames processed by the model per second.

\begin{table*}[ht]
\vspace{-0.5cm}
  \centering
  \renewcommand{\arraystretch}{1.4}
  \begin{tabular}{ccccccccccc}
  \toprule
    \multirow{2}{*}{\textbf{Train Dataset}} & \multicolumn{2}{c}{\textbf{OCR}} & \multicolumn{2}{c}{\textbf{MER}} & \multicolumn{2}{c}{\textbf{TR}} &{\textbf{DLA}} \\
    \cmidrule(lr){2-3} \cmidrule(lr){4-5} \cmidrule(lr){6-7} \cmidrule(lr){8-8}
    & {BLEU$\uparrow$} & {EditDis$\downarrow$} & {CDM$\uparrow$} & {CSR$_{MER}$$\uparrow$} & {F1$\uparrow$} & {CSR$_{TR}$$\uparrow$}& {F1$\uparrow$} \\   
    \hline
    Task-Specific &96.7 &2.2   &98.5    &99.8  &91.2  &92.7  &87.8                      \\
    \hline
    OCR+DLA &96.1 &2.4   &-    &-  &-  &-  &\textbf{88.9}                      \\
    \hline
    OCR+MER+TR   &\textbf{97.1}    &\textbf{1.8}  &\textbf{98.9}  &\textbf{99.9}  &\textbf{92.3}      &\textbf{94.6}     &-    \\
    \toprule
  \end{tabular}
  \caption{\label{recognition}
    Ablation experiments on task collaboration, comparison of task performance when using \textbf{Task-specific} training, where each task is trained independently, and other joint multi-task strategies.
  }
\end{table*}

\subsection{Selection of Baseline Models}
For the MER task, we selected UniMERNet \cite{wang2024unimernetuniversalnetworkrealworld}, the current state-of-the-art (SOTA) model, and Texify \cite{paruchuri2023texify}, which has shown strong competitive performance in recent evaluations. In the OCR task, we compared several models, including the large-scale model TextMonkey \cite{liu2024textmonkeyocrfreelargemultimodal} and smaller models such as Nougat \cite{blecher2023nougatneuralopticalunderstanding}, for a multi-scale evaluation. For the TR task, we evaluated our approach against StructEqTable \cite{xia2024docgenome}, one of the most representative models in current Table-to-Sequence methods. In the DLA task, we compared our method with two major object detection frameworks, YOLO and DETR (e.g., DocLayout-YOLO \cite{zhao2024doclayoutyoloenhancingdocumentlayout}, Deformable-DETR \cite{https://doi.org/10.48550/arxiv.2010.04159}). Although GOT \cite{wei2024generalocrtheoryocr20} is not capable of handling the DLA task, it performs well in the other three recognition tasks, making it a relevant model for comparison.

\subsection{Implementation Details}
We conducted our experiments using the PyTorch framework on eight NVIDIA H100 GPUs, with an initial learning rate of 1e-5, a per-GPU batch size of 12, and employing a cosine learning rate scheduler to progressively adjust the model parameters.

\subsection{Main Results}

\subsubsection{MER performance}

We use the open-source UniMER-1M \cite{wang2024unimernetuniversalnetworkrealworld} as the evaluation dataset to assess the performance on MER. Since DocFusion is specifically designed for processing printed documents, the primary evaluation focuses on the Simple Printed Expression (SPE) and Complex Printed Expression (CPE) subsets of UniMER-1M. As shown in Table~\ref{main}, DocFusion performs exceptionally well across multiple evaluation metrics, particularly in CSR and ExpRate. Notably, its ExpRate surpasses the second-ranked UniMERNet by 5.2\%, demonstrating superior reliability in real-world document parsing. The results presented here merge the performance of both SPE and CPE, with detailed separate results and handwritten expressions provided in \ref{specpehwe}.

\subsubsection{TR performance}
We selected DocGenome \cite{xia2024docgenome} as the evaluation dataset because it offers a comprehensive collection of 500K scientific documents across various disciplines, covering a wide range of document-oriented tasks. From this dataset, we extracted 3,000 LaTeX-based table samples as the test set. Using LatexNodes2Text, we extracted the content of each table cell to compute F1 scores. As shown in Table~\ref{main}, DocFusion excels on this benchmark, achieving F1 scores that surpass those of the second-ranked model by 1.6\%, while having less than one-third of its parameter count. \textbf{\textit{Note:}} In this work, in order to maintain consistency with MER and explore multi-task collaboration, we also chose Latex as the output format for our TR task. However, in the past, Latex was not mainstream in Table-to-Sequence tasks, so there are fewer models available for comparison. To provide more comprehensive reference information, we have included the F1 scores of other models that output in HTML in the appendix \ref{TR}.

\subsubsection{OCR performance}
As mentioned in \ref{training datasets}, DocLayNet \cite{doclaynet2022} supports not only DLA but also OCR evaluation. We selected 3,000 English image samples from the dataset as the test set. As shown in Table~\ref{main}, DocFusion achieves exceptional performance in both BLEU and EditDis. This outstanding result is primarily attributed to DocFusion's joint training on three recognition tasks, which enhances its efficiency and effectiveness in handling complex document structures.

\subsubsection{DLA performance}
We use the test set from DocLayNet\cite{doclaynet2022} to evaluate the DLA task. In terms of FPS, while DocFusion exhibits a slight disadvantage in processing speed, it offers an out-of-the-box solution that eliminates the need for hyperparameter tuning in practical applications. This enables the model to achieve optimal performance directly, without requiring further adjustments, thereby compensating for its lower speed.

Regarding accuracy, DocFusion generates layout element labels and coordinates by sequentially predicting tokens without relying on confidence scores. Given that the commonly used Average Precision (AP) metric in object detection is based on confidence scores, it is not directly applicable in this evaluation. To ensure a fair comparison with confidence-based models, we adopt an alternative evaluation methodology. Specifically, for these models, we compute Precision, Recall, and F1-score at various thresholds and select the maximum F1-score across all thresholds as the final evaluation metric. As shown in Table~\ref{end_dla}, DocFusion demonstrates strong performance in the domain of scientific document detection.

\begin{table*}[ht]
\vspace{-0.5cm}
  \centering
  \renewcommand{\arraystretch}{1.4}
  \begin{tabular}{lcccccccr}
  \toprule
    \multirow{2}{*}{\shortstack{\textbf{Objective}\\\textbf{Function}}} & \multicolumn{2}{c}{\textbf{OCR}} & \multicolumn{2}{c}{\textbf{MER}} & \multicolumn{2}{c}{\textbf{TR}} & \multicolumn{2}{c}{\textbf{DLA}} \\

    \cmidrule(lr){2-3} \cmidrule(lr){4-5} \cmidrule(lr){6-7} \cmidrule(lr){8-9} 
     & {BLEU$\uparrow$} & {EditDis$\downarrow$} & {CDM$\uparrow$} & {CSR$_{MER}$$\uparrow$} & {F1$\uparrow$} & {CSR$_{TR}$$\uparrow$} & {F1$\uparrow$} \\   
    \hline
    % Base (CE) &99.1 &0.008   &98.9    &99.9  &94.3  &96.6 &-    \\
    % \hline
    CE &96.5  &2.3  &97.8    &96.5  &90.2  &89.1 & 87.9 \\

    GK-CEL &\textbf{97.4}    &\textbf{1.8}  &\textbf{98.7}    &\textbf{99.8}  &\textbf{92.1}  &\textbf{92.5} & \textbf{88.4}\\
    \toprule
  \end{tabular}
  \caption{\label{objective_func}
    Ablation analysis of Gaussian-Kernel Cross-Entropy Loss was conducted on the same dataset across four tasks: OCR, MER, TR, and DLA. \textbf{CE} represents training with the standard cross-entropy loss, while \textbf{GK-CEL} denotes training with Gaussian-Kernel Cross-Entropy Loss.}
\end{table*}

\subsection{Ablation Study}

\subsubsection{OCR-Driven Enhancement of DLA}

This section explores the impact of OCR on DLA performance. As shown in Table \ref{recognition}, the results in the DLA column from the first and second rows indicate that adding the OCR task improves DLA performance, with an F1 increase of up to 1.3\%. This result demonstrates the effectiveness of using textual information in joint training. Compared to independent training that relies only on visual features, OCR significantly enhances the model's robustness. For example, tables and mathematical expressions have distinct visual features, which can often be effectively recognized by DLA models. In contrast, text or titles have less distinctive visual features, making it challenging to predict their labels based on visual information alone. By providing complementary textual information, OCR strengthens the collaboration between visual and semantic features, resulting in better overall performance.

\subsubsection{Collaboration of Recognition Tasks}

In this section, we explore the collaboration among the recognition tasks OCR, TR, and MER. As shown in Table \ref{recognition}, the experimental results from the first and third rows demonstrate that joint training yields better performance compared to training each task individually. Specifically, OCR achieves a 0.4\% improvement in BLEU score, MER sees increases of 0.4\% in CDM and 0.1\% in CSR, and TR benefits most significantly, with a 1.2\% improvement in F1 score for cell parsing and a 2.1\% increase in CSR. This collaboration enables the model to leverage shared information across tasks, enhancing individual task performance and improving overall document parsing capabilities. These results demonstrate that multi-task collaboration effectively enhances performance by leveraging shared information.

\begin{figure}[t]
  \includegraphics[width=\columnwidth, trim=260 95 250 102, clip]{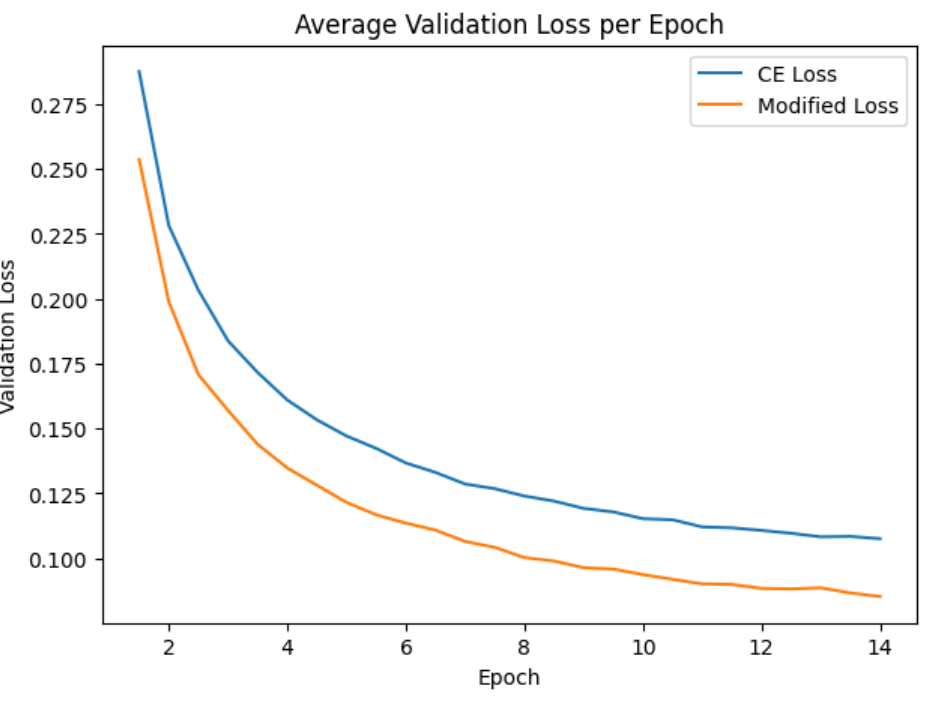}
  \caption{Validation loss curves under identical hyperparameter settings, where the only variation is the choice of the objective function. }
  \label{fig:experiments}
\end{figure}

\subsubsection{Results of improved objective function}

In this section, we compared the original cross-entropy and Gaussian-Kernel Cross-Entropy Loss (Gk-CEL) in recognition and detection tasks. As shown in Table \ref{objective_func}, the results demonstrate that Gk-CEL led to significant performance gains across both task categories. In recognition tasks, the BLEU score in the OCR task saw an improvement of 0.9\%. Additionally, the CDM metric in the MER task increased by 0.9\%, while the F1 score in the TR task rose by 2.1\%. Notably, for the CSR metric, which assesses LaTeX compilation success, the MER and TR tasks achieved gains of 3.4\% and 3.8\%, respectively, highlighting enhanced usability and correctness of the LaTeX outputs. For the detection task, the F1 score of the DLA task increased by 0.5\%. This improvement can be attributed to Gk-CEL, which alleviates the issue of coordinate token errors dominating the gradient. By addressing this imbalance, the objective function not only enhances the performance of recognition tasks but also improves the accuracy of predicting layout element categories in the detection task itself. These results collectively show that Gk-CEL effectively addresses key challenges in loss minimization, ensuring that tasks such as DLA can operate within a generative framework. It avoids gradient dominance issues while achieving better task balance in a multi-task learning setup.

\section{Conclusion}
In this work, we introduced DocFusion, the first approach to integrate the four modules of a document parsing pipeline into a unified model by designing Gaussian-Kernel Cross-Entropy Loss tailored to handle diverse data types across tasks. Our method achieved SOTA performance on multiple benchmarks. To enable downstream applications, we re-annotated the widely used DocLayNet dataset and constructed a large-scale formula-to-LaTeX dataset, applying a unified standardization process. Through detailed analysis, we observed that DocFusion, as a lightweight model, effectively integrates multiple tasks into a single framework, demonstrating both efficiency and versatility in handling complex document parsing challenges. In the future, we aim to extend DocFusion to larger models and further improve dataset standardization to enhance its performance and applicability across broader tasks and domains.

\section*{Limitations}
While this study primarily focuses on three recognition tasks using standard PDF screenshots, we have enhanced the model's generalization capabilities by incorporating handwritten mathematical expressions. However, the model still has limitations in handling handwritten or other non-standard table formats. For the detection task, although the model demonstrates competitive performance in both accuracy and usability, its processing speed presents challenges for real-time or high-throughput applications. This highlights the need for further optimization in computational efficiency to better meet diverse application demands.

\section*{Acknowledgements}
TThe authors wish to thank the anonymous reviewers for their helpful comments. This work was funded by National Natural Science Foundation of China (No.62476061)

\bibliography{custom}

\newpage
\appendix

\section{Details of Datasets}

\subsection{DLA Dataset Reconstruction}
\begin{figure}[h]
  \includegraphics[width=\columnwidth]{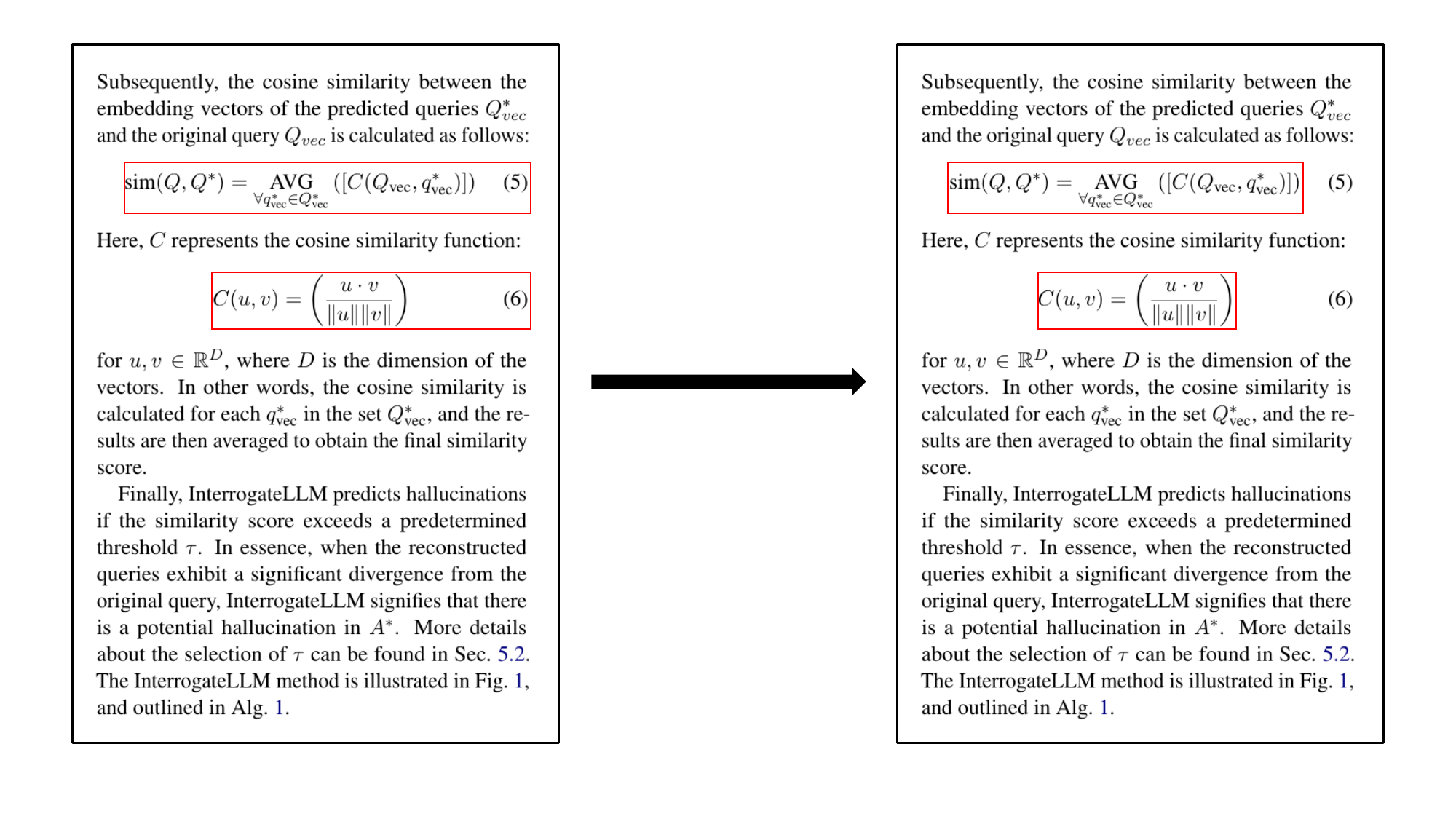}
  \caption{The corresponding numbers were removed from the annotated data for mathematical expression detection.}
  \label{YOLO}
\end{figure}
In DocLaynet and other similar datasets, the annotation of mathematical formulas has certain limitations, as show in figure \ref{YOLO}, the content of math expression and numbering are typically annotated within the same bounding box. This annotation approach introduces noise in subsequent Mathematical Expression Recognition (MER) tasks. 

To address this issue, we extracted formulas from arXiv LaTeX source files using regular expressions and assigned unique colors and bounding boxes to each element. Then, we employed a fuzzy matching algorithm to ensure annotation accuracy and eliminate overlaps. Finally, we trained a lightweight detection model and, combined with manual verification, re-annotated pages containing formulas. These improvements significantly enhance the dataset's applicability to subsequent MER tasks.

\subsection{MER and TR Dataset}
\textbf{MER Dataset.} The UniMER-1M \cite{wang2024unimernetuniversalnetworkrealworld} has significantly advanced MER research but contains many redundant spaces in LaTeX code. Although some spaces are syntactically necessary, most are unnecessary, increasing output length and computational overhead. To address this, we constructed a new dataset by extracting content from LaTeX files, normalizing style variations and verifying accuracy through re-rendering. Models trained on our dataset produce LaTeX code that is approximately 34.2\% shorter for complex expressions and 37.5\% shorter for simple expressions on the UniMER-1M test set, demonstrating improved efficiency.\\

\textbf{TR Dataset.} In the TR task of DocFusion, we adopted LaTeX as the output format for two main reasons: (1) to ensure consistency with the MER task’s output format, enabling better multi-task collaboration; and (2) because LaTeX facilitates both the extraction of cell content and the restoration of the original table layout. Existing LaTeX-based TR datasets either lack sufficient scale or fail to separate tables from captions, conflicting with our DLA task. To overcome these limitations, we constructed a high-quality TR dataset with 100K samples by following a similar approach to the MER dataset.\\

\subsection{Latex-based data standardization}
\label{appendixMERandTR}

\begin{table}[h]
  \centering
  \begin{tabular}{lcc}
    \toprule
    \textbf{Issue} & \textbf{Original} & \textbf{Standardized} \\
    \hline
   % Compare & \texttt{\textbackslash geq} & \texttt{\textbackslash ge} \\
   Bracket & \texttt{\textbackslash \{} & \texttt{\textbackslash lbrace} \\
   Subsup &\texttt{a\textasciicircum 1\_2} & \texttt{a\_2\textasciicircum 1} \\
    Prime                 & a$'$ & \texttt{a\textasciicircum{\{\textbackslash prime\}}}\\
    Fraction     & \texttt{\textbackslash over} & \texttt{\textbackslash frac} \\ 
    Space       & \texttt{\textbackslash tabular\{l c\}} & \texttt{\textbackslash tabular\{lc\}}\\
    \toprule
  \end{tabular}
  \caption{\label{stand}
    Examples of LaTeX standardization for various symbols and expressions.
  }
\end{table}

We chose to standardize the output format as LaTeX for two recognition tasks involving non-plain-text elements. For MER, converting to LaTeX was essential as it provides a precise representation of mathematical formulas. For TR, in addition to ensuring format consistency, converting to LaTeX also allows for the restoration of the original content through compilation, and enables the extraction of cell elements using tools such as LatexNodes2Text, thus enhancing processing flexibility. 

We used regular expressions to extract relevant content from the LaTeX source files of research papers. However, due to variations in author writing styles, the same formula or table may appear in multiple forms, increasing the complexity of training. As show in table \ref{stand} , we analyzed these different representations, standardized them to eliminate ambiguities and ensured consistency. To verify the accuracy of the standardized LaTeX code, we re-rendered it into images, creating a high-quality dataset that aligns with the actual input-output content.

\begin{table*}[h]
  \centering
  \setlength{\tabcolsep}{5.7pt}
  \resizebox{\textwidth}{!}{
  % \small
  \begin{tabular}{lcccccccccc}
  \toprule
    \multirow[c]{2}{*}{\textbf{Model}} & \multirow[c]{2}{*}{\textbf{size}}&\multicolumn{3}{c}{\textbf{SPE}} & \multicolumn{3}{c}{\textbf{CPE}}&\multicolumn{3}{c}{\textbf{HWE}}  \\
    \cmidrule(lr){3-5} \cmidrule(lr){6-8} \cmidrule(lr){9-11}
& &CDM$\uparrow$&ExpRate$\uparrow$ &CSR$\uparrow$ &CDM$\uparrow$&ExpRate$\uparrow$ &CSR$\uparrow$ &CDM$\uparrow$&ExpRate$\uparrow$ &CSR$\uparrow$\\
    \hline
    Pix2tex \citeyearpar{blecher2022pix2tex} &- &92.1 &59.0    &99.8  &45.2  &7.2  &88.1  &24.7  &8.1 &16.3   \\
    Texify \citeyearpar{paruchuri2023texify} &312M &98.7 &89.8    &99.8  &69.8  &35.6  &94.3  &49.9  &21.3 &25.8   \\
    GOT \citeyearpar{wei2024generalocrtheoryocr20} &580M  &95.0 &82.7    &98.6  &73.3  &36.4  &96.4  &31.2  &17.7 &10.2   \\
    UniMERNet \citeyearpar{wang2024unimernetuniversalnetworkrealworld} &325M  &\textbf{99.7} &95.6    &\textbf{99.9}  &\textbf{97.6}  &77.4  &99.2  &\textbf{94.7}  &65.3 &98.1\\
    DocFusion(Ours)  &289M   &\textbf{99.7} &\textbf{97.3}    &\textbf{99.9}  &96.9  &\textbf{88.1}  &\textbf{99.5}  &94.1  &\textbf{72.1} &\textbf{99.3}\\
    \toprule
  \end{tabular}
  }
  \caption{\label{specpehwetable}
    Supplementary details of \textbf{MER}. \textbf{SPE} refers to simple printed mathematical expressions, \textbf{CPE} refers to complex printed mathematical expressions, and \textbf{HWE} refers to handwritten mathematical expressions.
  }
\end{table*}

\section{Other supplementary experiments}

\subsection{Details of MER Performance}
\label{specpehwe}
we provide a detailed presentation of the main experimental results for MER, showing the performance of the relevant models on simple, complex, and non-standard handwritten mathematical expressions. For specifics, please refer to Table \ref{specpehwetable}.

% \newpage
\subsection{Other Table-to-Sequence Method}
\label{TR}

\begin{table}[h]
  \centering
  \begin{tabular}{lcc}
    \toprule
    \textbf{Methods} & \textbf{F1} & \textbf{CSR} \\
    \hline
   % Compare & \texttt{\textbackslash geq} & \texttt{\textbackslash ge} \\
   surya & 37.4 & - \\
   ppstructure\_table& 78.1 & - \\
   Deepdoctection& 53.7 & - \\
   RapidTable& 87.9 & - \\
   MixTex& 46.2 & 27.4 \\
   GOT& 86.9 & 81.6 \\
   StructEqTable& 90.6 & 93.2 \\
   DocFusion& 92.1 & 92.5 \\
    \toprule
  \end{tabular}
  \caption{\label{trtable}
    Due to differences in the method of extracting cell contents, the fairness of the experiment cannot be guaranteed, therefore, it is provided for reference only.
  }
\end{table}

This study aims to explore multi-task collaboration, and therefore, the TR task also adopts Latex as the output format to maintain consistency with MER. However, Latex has not been the mainstream approach for TR tasks in recent times, resulting in a limited number of TR models available for comparison in the main experiment. To address this limitation, we incorporated other methods based on HTML as the output format. \textbf{However, due to differences in sequence extraction methods, ensuring a fair comparison is challenging. Therefore, we have included the supplementary experimental results in the appendix for reference.}

% \cleardoublepage
\section{Other optimization methods}
The challenge of this experiment lies in effectively optimizing continuous coordinate-type data within a discrete generative framework. Since there are inherent errors in coordinate annotations, these errors are further amplified when training the generative framework using cross-entropy loss, especially when the framework performs multiple tasks, which exacerbates the issue. To address this problem, in addition to the Gaussian-Kernel Cross-Entropy Loss introduced in the main text, we employed several other optimization strategies, including the basic adjustments of data ratios or loss weights, as well as using soft-argmax to continuously map discrete coordinate tokens.

\subsection{Hyperparameters Adjustment Strategies}
The root cause of the training difficulty lies in the fact that the discrete coordinate tokens do not effectively dominate the loss during training, leading to poor gradient propagation and inefficient parameter updates. To address this, one possible solution is to adjust the data ratios or the loss weights across different task types. However, while this approach can improve training stability to some extent, it is overly engineering-driven and does not fundamentally solve the underlying issue of inadequate gradient flow caused by the discrete nature of the coordinate tokens.

\begin{table*}[ht]
\vspace{-0.5cm}
  \centering
  \renewcommand{\arraystretch}{1.4}
  \begin{tabular}{lccccccccr}
  \toprule
    \multirow{2}{*}{\shortstack{\textbf{Model}}} &\multirow{2}{*}{\shortstack{\textbf{size}}}& \multicolumn{2}{c}{\textbf{OCR}} & \multicolumn{2}{c}{\textbf{MER}} & \multicolumn{2}{c}{\textbf{TR}} & \multicolumn{2}{c}{\textbf{DLA}} \\

    \cmidrule(lr){3-4} \cmidrule(lr){5-6} \cmidrule(lr){7-8} \cmidrule(lr){9-10} &
     & {BLEU$\uparrow$} & {EditDis$\downarrow$} & {CDM$\uparrow$} & {CSR$_{MER}$$\uparrow$} & {F1$\uparrow$} & {CSR$_{TR}$$\uparrow$} & {F1$\uparrow$} \\   
    \hline
    % Base (CE) &99.1 &0.008   &98.9    &99.9  &94.3  &96.6 &-    \\
    % \hline

    DocFusion-base &289 &\textbf{97.4}    &\textbf{1.8}  &98.7    &\textbf{99.8}  &92.1  &\textbf{92.5} & \textbf{88.4}\\

    DocFusion-large &738 &97.2    &1.9  &\textbf{99.1}    &\textbf{99.8}  &\textbf{92.4}  &\textbf{92.5} & \textbf{89.1}\\
    \toprule
  \end{tabular}
  \caption{\label{size}
    Ablation analysis of Gaussian-Kernel Cross-Entropy Loss was conducted on the same dataset across four tasks: OCR, MER, TR, and DLA. \textbf{CE} represents training with the standard cross-entropy loss, while \textbf{GK-CEL} denotes training with Gaussian-Kernel Cross-Entropy Loss.}
\end{table*}

\subsection{Soft-argmax Strategies}
\begin{figure}[h]
  \includegraphics[width=\columnwidth]{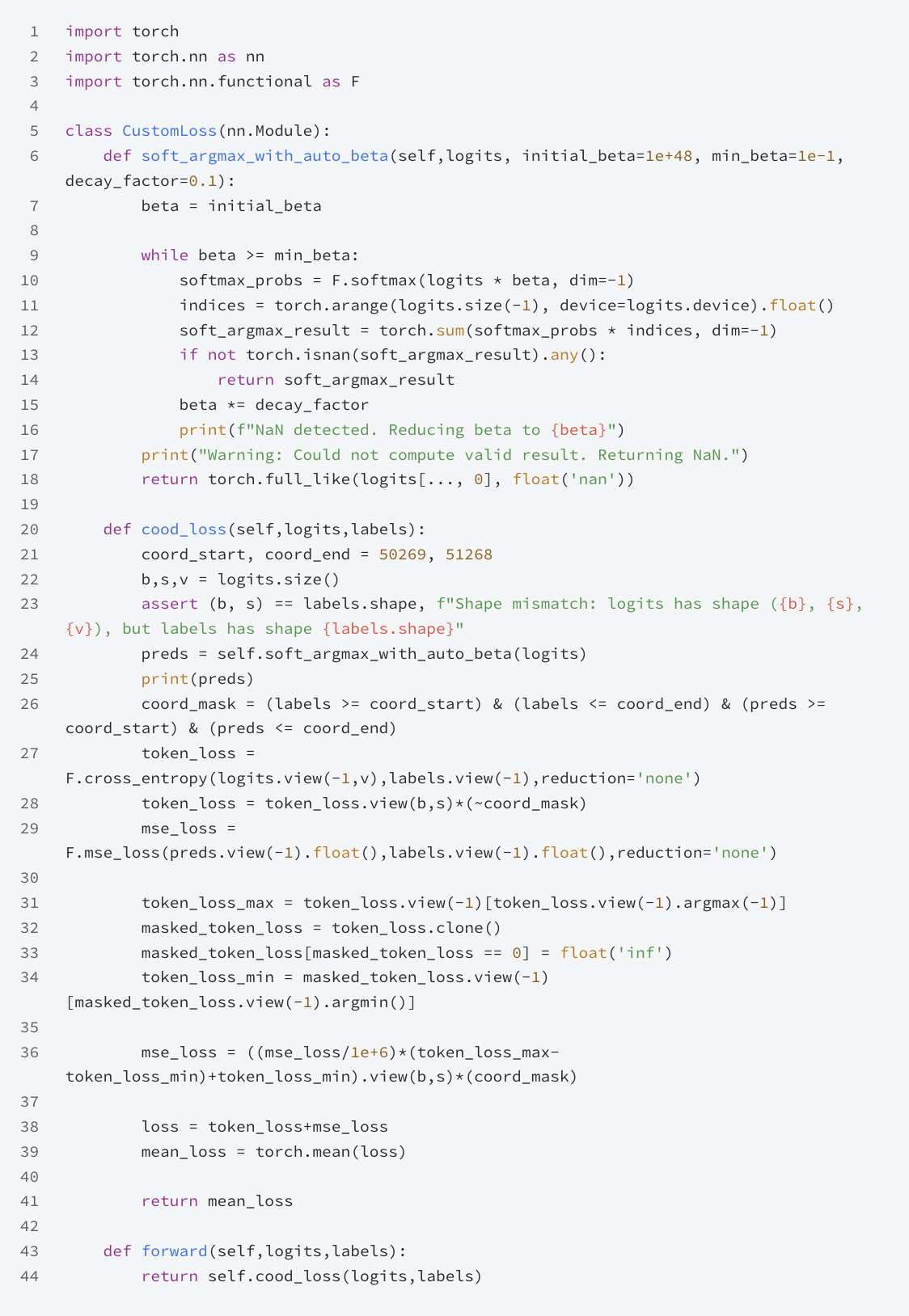}
  \caption{Soft-argmax Loss}
  \label{Soft-argmax Loss}
\end{figure}
The core issue lies in the fact that while multi-task frameworks need to be discrete, coordinates are inherently continuous. A natural solution to this problem is to "smooth" the coordinate loss, effectively making it continuous. This approach offers an intuitive way to handle the challenge, and we primarily use the soft-argmax technique to obtain the position coordinates while maintaining the gradient flow, followed by the computation of the loss via Mean Squared Error (MSE).

However, the difficulty arises during multi-task training: after calculating the MSE, we need to ensure that it remains within the same range as other cross-entropy (CE) losses. The challenge here is to maintain balance and prevent the MSE loss from overwhelming the CE losses. Moreover, if the hyperparameters of the soft-argmax are not set appropriately, it can easily lead to gradient explosion during training, further complicating the optimization process.

Although this method aims to address the issue at its core by making the coordinate loss continuous, it still relies heavily on the correct setting of hyperparameters. Furthermore, it presents generalization issues when applied to different tasks or datasets. In comparison, the Gaussian-Kernel Cross-Entropy Loss (GK-CEL) offers a more robust solution, as it reduces the dependency on hyperparameters while improving generalization performance.

\subsection{Model Size Analysis}
While model performance generally benefits from increased parameter size, the advantages can diminish in recognition-oriented tasks due to the limited gains in accuracy relative to the added computational cost. In the early stages of this work, we trained a larger version of our model with 738M parameters. Although it achieved slightly better performance on certain tasks—such as a modest improvement on the DLA benchmark—the gains were not substantial enough to justify the significantly higher inference cost, especially given the autoregressive nature of decoding.

As our primary goal is to demonstrate the feasibility of a lightweight and unified model, we chose to focus on the 289M version of DocFusion in this paper. As shown in Table\ref{size}, this smaller model already delivers strong results across tasks, including 97.4 BLEU on OCR and 99.8 CSR on MER. We believe this better reflects the practical trade-off between efficiency and performance. Results from the larger variant will be included in a future version to facilitate further exploration and provide a reference for the research community.

\cleardoublepage
\begin{figure*}[h]
  \includegraphics[width=0.96\linewidth]{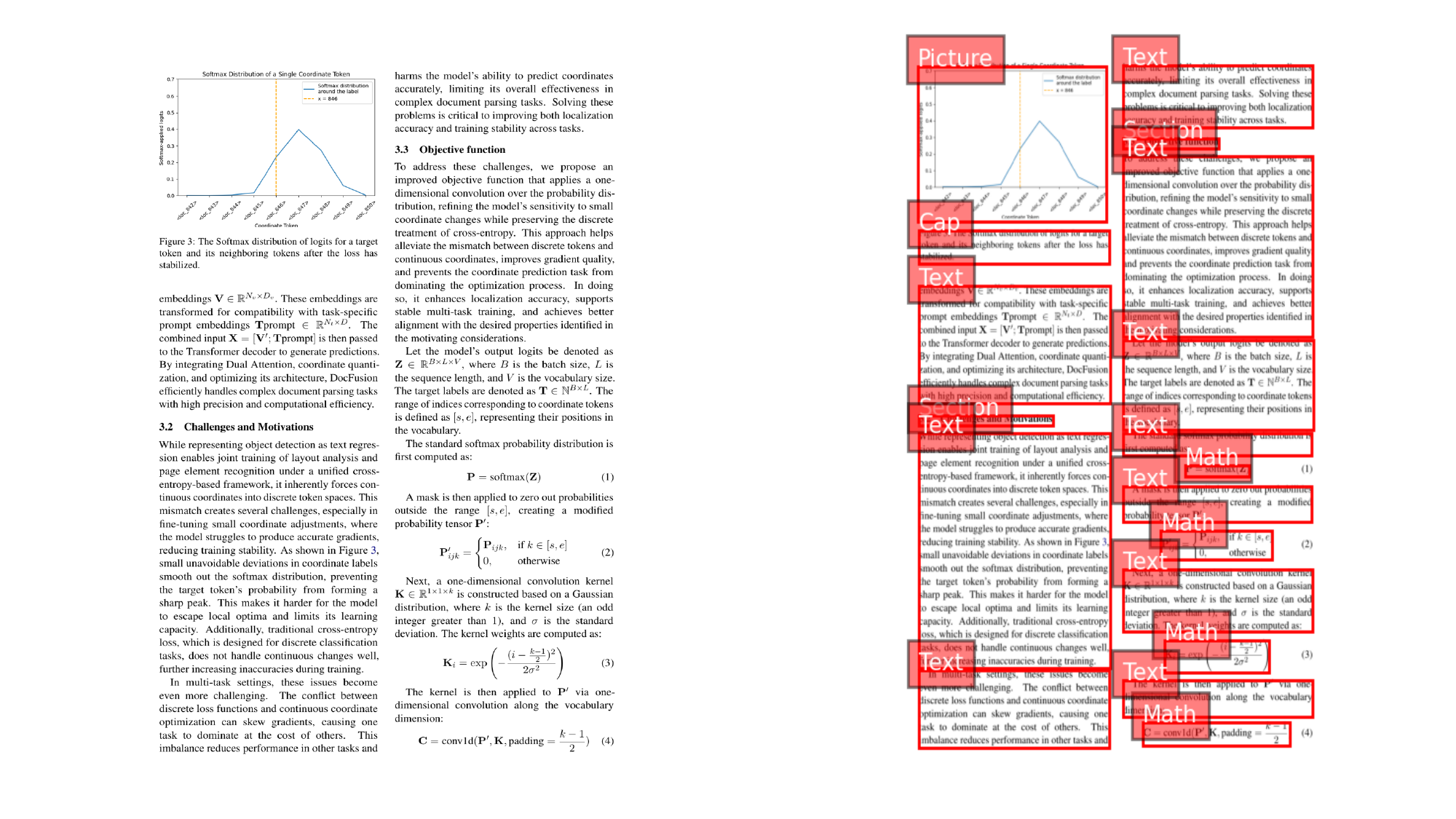}
  \caption {DLA Effect Presentation}
\end{figure*}

\cleardoublepage
\begin{figure*}[h]
  \includegraphics[width=0.96\linewidth]{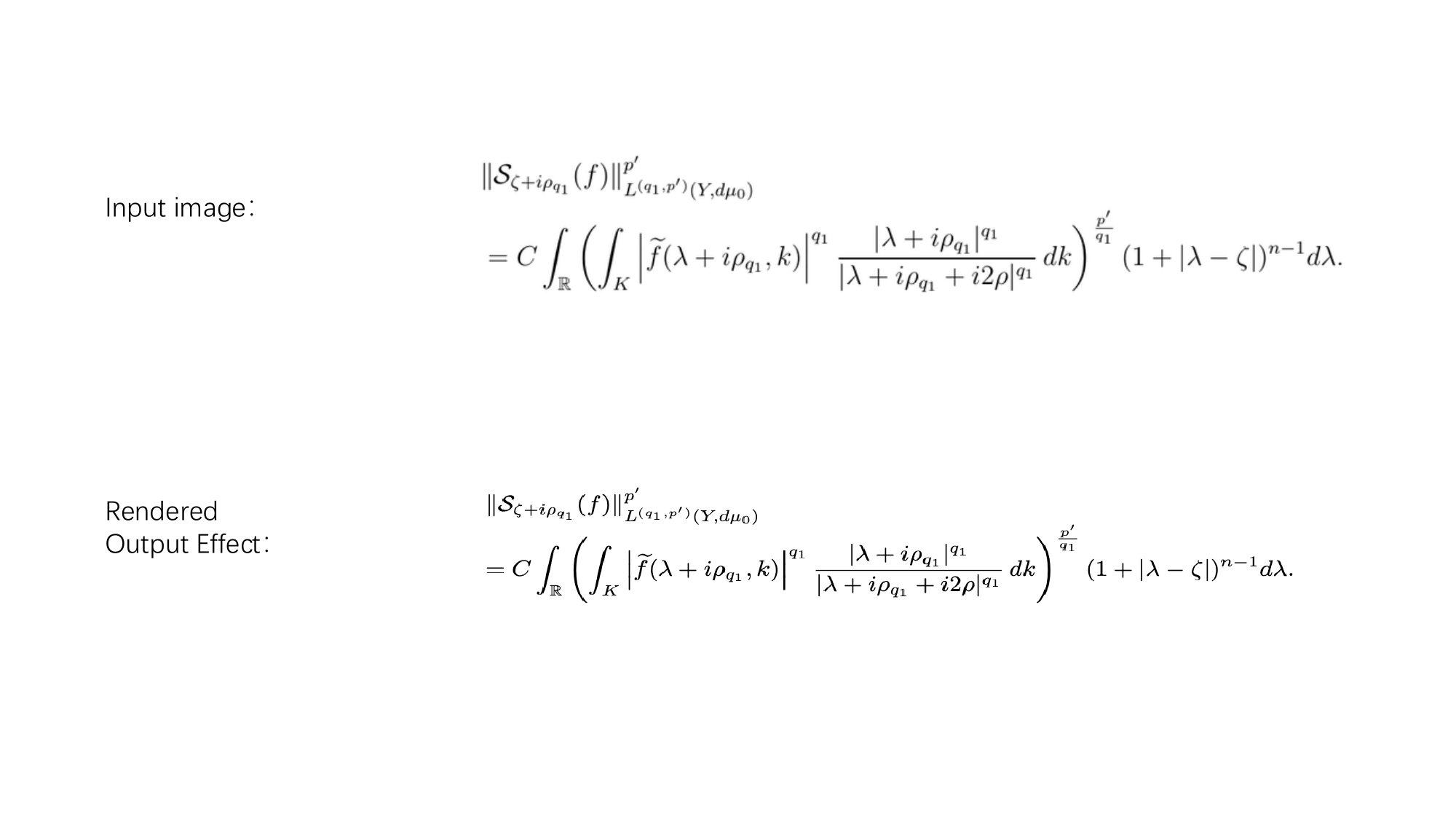}
  \caption {MER Effect Presentation}
\end{figure*}

\cleardoublepage
\begin{figure*}[h]
  \includegraphics[width=0.96\linewidth]{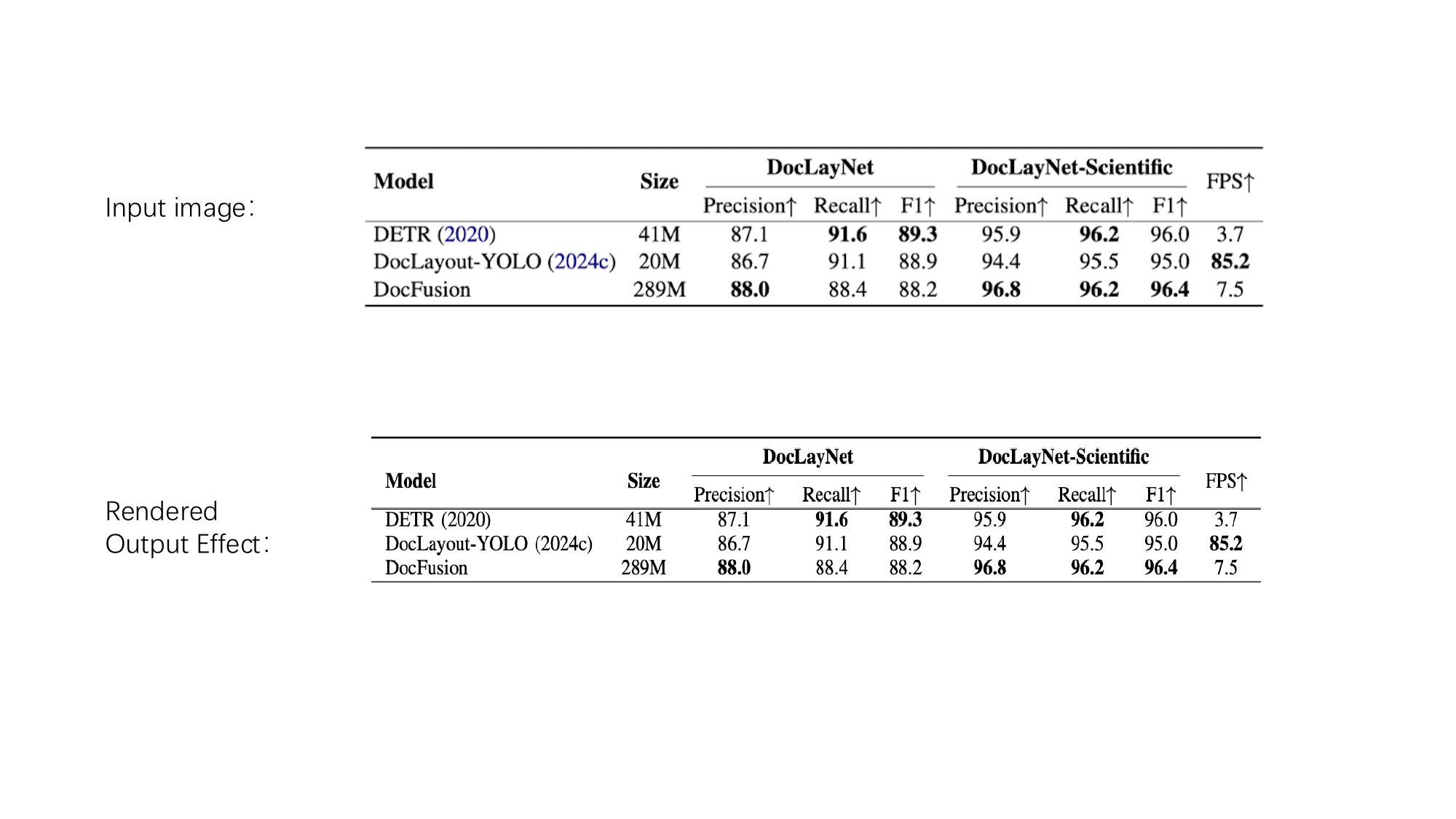}
  \caption {TR Effect Presentation}
\end{figure*}
\end{document}